\documentclass[runningheads]{llncs}
\usepackage[T1]{fontenc}
\usepackage{amsmath}
\usepackage{graphicx}
\usepackage{hyperref}
\usepackage{amssymb}
\begin{document}
\title{Patch-based Automatic Rosacea Detection Using the ResNet Deep Learning Framework}
\author{Chengyu Yang\text \and Rishik Reddy Yesgari\and Chengjun Liu}
\authorrunning{Chengyu et al.} % abbreviated author list (for running head)
% \tocauthor{Ivar Ekeland, Roger Temam, Jeffrey Dean, David Grove,
% Craig Chambers, Kim B. Bruce, and Elisa Bertino}
%
\institute{New Jersey Institute of Technology, Newark NJ 07103, USA,\\
\email{\{cy322, ry248, chengjun.liu\}@njit.edu},
% \and New Jersey Institute of Technology, Newark NJ 07103, USA,\\
% \email{cliu@njit.edu},
\\
WWW home page:\\
\url{https://chengyuyang-njit.github.io/}\\
\url{https://rishik-portfolio-chi.vercel.app/}\\
\url{https://web.njit.edu/~cliu/}
}

\maketitle              % typeset the header of the contribution
\begin{abstract}
Rosacea, which is a chronic inflammatory skin condition that manifests with facial redness, papules, and visible blood vessels, often requirs precise and early detection for significantly improving treatment effectiveness. 
This paper presents new patch-based automatic rosacea detection strategies using the ResNet-18 deep learning framework. The contributions of the proposed strategies come from the following aspects. First, various image pateches are extracted from the facial images of people in different sizes, shapes, and locations. Second, a number of investigation studies are carried out to evaluate how the localized visual information influences the deep learing model performance. Third, thorough experiments are implemented to reveal that several patch-based automatic rosacea detection strategies achieve competitive or superior accuracy and sensitivity than the full-image based methods. And finally, the proposed patch-based strategies, which use only localized patches, inherently preserve patient privacy by excluding any identifiable facial features from the data. 
The experimental results indicate that the proposed patch-based strategies guide the deep learning model to focus on clinically relevant regions, enhance robustness and interpretability, and protect patient privacy. As a result, the proposed strategies offer practical insights for improving automated dermatological diagnostics.

\end{abstract}
\section{Introduction}
Rosacea is a chronic inflammatory skin disorder characterized by persistent facial redness, papules, pustules, and visible blood vessels\cite{ref1}. It affects millions worldwide, with a higher prevalence among fair-skinned individuals, and can significantly impact quality of life if left untreated. Accurate and early detection is critical for effective management, as timely intervention can prevent progression and alleviate symptoms. In recent years, deep learning–based image analysis has shown great promise in dermatological diagnosis, offering automated and objective assessments from facial photographs. However, many existing methods rely on full-face images, which not only increase computational burden but also raise privacy concerns due to the inclusion of personally identifiable features.

In dermatology, the visual cues essential for diagnosis are often localized to specific facial regions, such as the forehead, cheeks, and nose\cite{ref2}. This suggests that analyzing smaller, targeted patches rather than the full face could guide models to focus on clinically relevant regions. Patch-based learning may also improve robustness by reducing the influence of irrelevant background information and by allowing the model to learn region-specific features more effectively. Moreover, by limiting the input to partial facial regions, such approaches inherently preserve patient privacy, making them useful for privacy-sensitive medical applications.

While patch-based methods have been explored in related fields such as histopathology\cite{ref3} and retinal imaging\cite{ref4}, their systematic evaluation for rosacea detection remains underexplored. Key questions include whether patch-based classification can match or exceed the performance of full-image models, and how different patch sizes and locations influence diagnostic accuracy and sensitivity. Furthermore, understanding the trade-offs between performance, interpretability, and privacy is essential for deploying such systems in real-world clinical settings.

In this study, we present a comparative analysis of various patching strategies for rosacea detection using a ResNet-18\cite{ref5} based classification framework. We investigate how patches extracted from different facial regions and with varying configurations affect model performance. Our results show that while not all patch configurations outperform full-image inputs, several achieve competitive or superior accuracy and sensitivity. Importantly, all patch-based approaches inherently safeguard patient privacy by excluding identifiable facial features. This work highlights the potential of region-focused learning to balance diagnostic accuracy, interpretability, and privacy protection, offering practical guidance for the design of automated dermatological diagnostic tools.

\section{Related Work}

Deep learning has significantly advanced dermatological image analysis, with convolutional neural networks (CNNs) achieving dermatologist-level performance in lesion classification tasks. Esteva et al. \cite{ref11} demonstrated the feasibility of CNNs for skin cancer diagnosis, while Tschandl et al. \cite{ref12} introduced the HAM10000 dataset, enabling standardized evaluation across multiple skin conditions. Beyond dermatology, privacy-preserving image analysis has been studied in biometric and healthcare domains, where approaches such as facial obfuscation, blurring, and masking aim to reduce re-identification risks \cite{ref13}. In medical contexts, federated learning and differential privacy have also been explored \cite{ref14}, though these often degrade diagnostic utility. Within dermatology, prior works have mostly relied on full-face or lesion-centric images, which both increase computational burden and raise concerns around patient identifiability.

Patch-based learning has shown strong potential in related fields. In computational pathology, Campanella et al. \cite{ref15} trained weakly supervised networks on millions of histopathology patches, demonstrating that regional analysis can outperform whole-slide evaluation. Wang et al. \cite{ref4} applied patch-based CNNs to retinal vessel segmentation, achieving high sensitivity to localized features. Similar region-focused strategies have been applied in chest X-rays and mammography, where focusing on smaller regions improves sensitivity and interpretability \cite{ref16}. Despite these successes, patch-based methods remain underexplored for rosacea, a condition where clinically relevant cues are often localized to central facial regions such as the cheeks, forehead, and nose. This gap motivates our systematic comparison of patching strategies for rosacea detection, where we examine how patch size, location, and configuration influence accuracy, interpretability, and privacy.

\section{Method}
\subsection{Patch Extraction}
To ensure consistent positioning of key facial features, reduces irrelevant background information, and facilitates the extraction of region-specific patches,
facial photographs are cropped to a uniform size and aligned based on the positions of eyes to standardize facial region localization. Four regions of interest (ROIs) were identified and selected: forehead (F), nose (N), left cheek (LC), and right cheek (RC).

\begin{figure}
    \centering
    \includegraphics[width=0.8\linewidth]{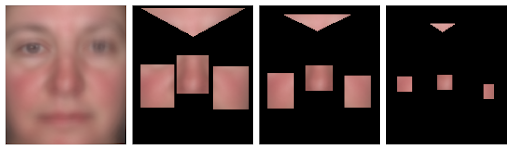}
    \caption{An example of patch selection with different sizes. From left to right: mean face, large size patches, medium size patches, small size patches}
    \label{fig_patch_extra}
\end{figure}

For each ROI, three patch scales are generated manually as is shown in Fig.\ref{fig_patch_extra}: large size patches, medium size patches and small size patches. The large size patches cover a broad area around the target region, including some surrounding skin. The medium size patches tightly center on the target region. The small size patches focus on the most clinically relevant central portion. 

\subsection{Patch Combinations}
\begin{figure}
    \centering
    \includegraphics[width=0.95\linewidth]{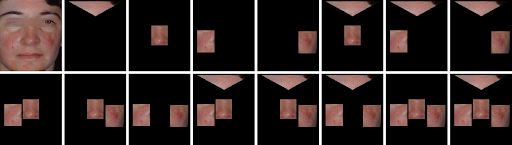}
    \caption{An example of different patch combinations for a face with rosacea. First row from left to right: full face, forehead (F) alone, the nose (N) alone, the left cheek (LC) alone, the right cheek (RC) alone, F+N, F+LC, F+RC. Second row from left to right: N+LC, N+RC, F+N+LC, F+N+RC, F+LC+RC, N+LC+RC and all four patches.}
    \label{fig1}
\end{figure}

We designed multiple experimental configurations to evaluate the contribution of different facial regions and categorized them according to the number of regions selected. Single-region patches include only one region, such as the forehead (F) alone, the nose (N) alone, the left cheek (LC) alone, or the right cheek (RC) alone. Two-region combinations consist of pairs of regions, including F+N, F+LC, F+RC, N+LC, and N+RC. Three-region combinations include F+N+LC, F+N+RC, F+LC+RC, and N+LC+RC. We also tested an all-region patch combining all four regions (F+N+LC+RC). For comparison, the full-face image was included as a baseline. An example is shown in Fig.\ref{fig1}. Unless otherwise stated, LC/RC in this paper are defined in the viewer frame: LC = image-left cheek (subject’s anatomical right), RC = image-right cheek (subject’s anatomical left). Each configuration was evaluated using large, medium, and small patch scales to investigate the effect of patch size on classification performance.

\subsection{Classification Model}
We used ResNet-18, a residual convolutional neural network with 18 layers that incorporates identity skip connections to improve gradient flow and enable deeper model training. The networks was initialized with ImageNet\cite{ref6} pretrained weights. The final fully connected layer was replaced with a single output neuron followed by a sigmoid activation to predict the probability of rosacea.

\section{Experiment}
In this section, we describe the experiments conducted to evaluate our proposed method. We first present the dataset used in our study and detail the preprocessing steps applied. Next, we outline the experimental setup and report the results.
\subsection{Dataset}
\begin{figure}
    \centering
    \includegraphics[width=0.95\linewidth]{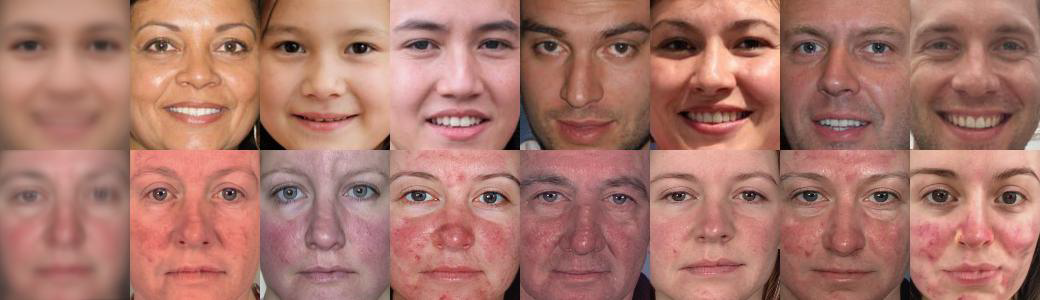}
    \caption{The first column shows the mean images of the two training datasets from the rosacea negative and positive classes, respectively. The remaining seven columns display seven pair of random example images from the two training datasets corresponding to the normal people and those with rosacea, respectively.}
    \label{fig2}
\end{figure}

As is shown in Fig.\ref{fig2}, since patients' data is scarce, private and hard to collect, we use the same datasets as \cite{ref7}\cite{ref8} that are synthetically generated from GANs. To improve data quality, these facial images are aligned and cropped to center the face. It is achieved by manually selecting the eyes' coordinates and aligned according to those coordinates.The main purpose is to remove background distractions and make mask application easier.
\subsubsection{Training and Validation} For rosacea positive cases, 300 frontal face images synthesized using a GAN framework \cite{ref9}\cite{ref10} are used, 250 of which are for training and 50 of which are for validation. For rosacea-negative cases, 600 facial images generated by StyleGAN\cite{ref17} are used, with 500 for training and 100 for validation. We applied the selected patch configurations to these images to assess the contribution of each facial region to detection performance.
\subsubsection{Testing} For real-world evaluation, we use 50 rosacea-positive images collected from Kaggle, DermNet, and the National Rosacea Society, all with at least 200 by 200 pixels in resolution. Each face is manually aligned, cropped and resized. We also include 150 rosacea-negative facial images from the CelebA dataset\cite{ref18} for testing. Test images were masked using the same patch configurations as those applied to the training and validation images.

\subsection{Deep Learning Model Configuration}

We fine-tune a ResNet-18 \cite{ref5} initialized with ImageNet-pretrained weights. The original fully connected layer at the end of ResNet-18 is replaced by a single-unit linear layer to produce a logit for binary classification:
\begin{equation}
z = f_{\theta}(\mathbf{x}), \qquad
\hat{y} = \sigma(z), \quad z \in \mathbb{R}.
\label{eq:model}
\end{equation}
Here, \(f_{\theta}\) denotes the fine-tuned ResNet-18 with parameters \(\theta\), and \(\sigma(\cdot)\) is the sigmoid function. All layers are updated end-to-end.

Inputs were fed at their aligned crop size of \(150\times130\times3\) (H\(\times\)W\(\times\)C; H=height, W=width, C=channels; C=3 for RGB). ResNet-18’s adaptive average pooling permits variable spatial dimensions. For patch-based experiments, the selected region-of-interest (ROI) mask(s) are applied to each image prior to normalization, yielding inputs that contain only the chosen ROI(s) (F, N, LC, RC) at one of three scales (small/medium/large).

Training uses binary cross-entropy with logits:
\begin{equation}
\mathcal{L}(\,z, y\,)= -\big[y\log \sigma(z) + (1-y)\log\big(1-\sigma(z)\big)\big]
\end{equation}

implemented via \texttt{BCEWithLogitsLoss}, which is numerically stable and fuses the sigmoid with the cross-entropy. Ground-truth labels are cast to shape $[N,1]$ to match the model output.

We use stochastic gradient descent with momentum:
\begin{equation}
 \theta \leftarrow \theta - \eta \,\nabla_{\theta}\mathcal{L} \quad\text{with}\quad \eta=10^{-3},\ \text{momentum}=0.9   
\end{equation}
A step decay schedule (\texttt{StepLR}) reduces the learning rate by a factor of $0.1$ every $7$ epochs (i.e., epochs 7, 14, 21, 28 for a 30-epoch run), yielding piecewise-constant LRs of $10^{-3}$, $10^{-4}$, $10^{-5}$, $10^{-6}$, and $10^{-7}$ toward the end. Mini-batch training is used for both training and validation iterators.

Each epoch alternates between \texttt{train} and \texttt{eval} phases. In the training phase we enable gradient computation, perform backpropagation, and update parameters; in validation we disable gradient computation and only forward-pass to compute loss and accuracy. After each epoch, the learning-rate scheduler is stepped. We checkpoint the model parameters with the highest validation accuracy observed during training; the final model restores this best checkpoint for downstream evaluation. At inference, probabilities are obtained via $\sigma(z)$. The configuration is summarized in Table\ref{tab:train_config}.

\begin{table}[t]
\centering
\caption{Training configuration used for all experiments.}
\label{tab:train_config}
\begin{tabular}{ll}
\hline
Base architecture & ResNet-18 (ImageNet-pretrained) \\
Classifier head & Linear(\texttt{num\_ftrs}, 1) \\
Loss & \texttt{BCEWithLogitsLoss} \\
Optimizer & SGD ($\eta=10^{-3}$, momentum $=0.9$) \\
LR schedule & StepLR (step\_size $=7$, $\gamma=0.1$) \\
Epochs & 30 (best checkpoint by val.\ accuracy) \\
Input size & $150\times130\times3$ (H$\times$W$\times$C) \\
\hline
\end{tabular}
\end{table}

\begin{figure}
    \centering    \includegraphics[width=1.0\linewidth]{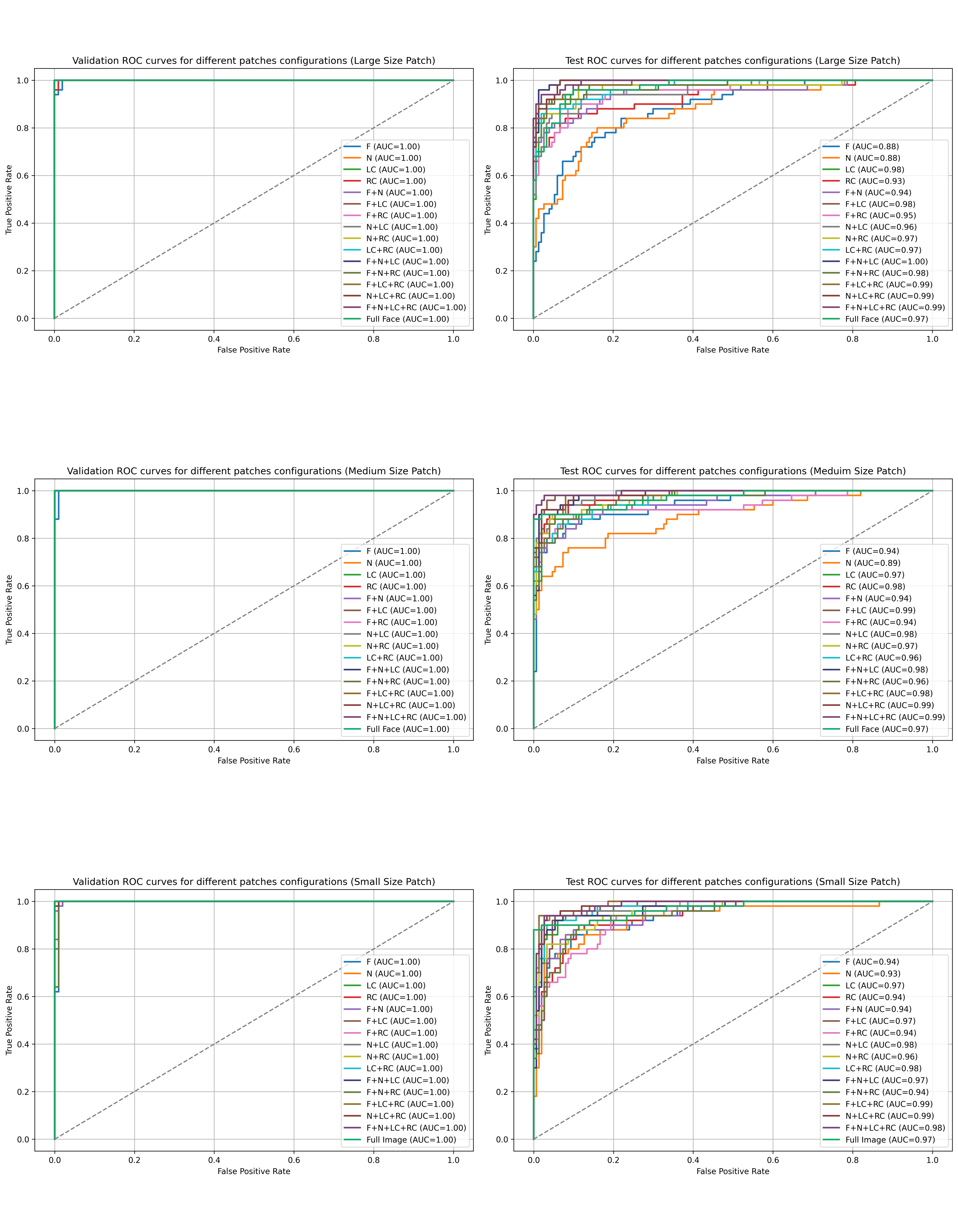}
    \caption{The ROC curves on validation and test set for different patch size configurations}
    \label{fig_results}
\end{figure}

\subsection{Results}
Fig.\ref{fig_results} shows the ROC curves on validation and test set for different patch size configurations and their corresponding area under the curve (AUC). The model performs well across different patch configurations on the validation set, suggesting that it has successfully learned to separate rosacea-positive and negative cases within the generated data distribution. However, this perfect validation performance is largely due to the high similarity between the training and validation sets, both of which consist of synthetic images, and therefore may not fully reflect true generalization ability. To test the model's generalization ability, we evaluate it on real-world test data. Unlike the validation results, the test ROC curves reveal noticeable performance gaps among different patch configurations.
\subsubsection{General Trends Across All Patch Sizes}
We can see that combination of patches consistently outperforms single patches. For all three patch sizes used, using multiple regions (e.g. F+N+LC or F+N+LC+RC) yields the highest AUC, often reaching 0.99 - 1.00. This shows that different facial regions provide complementary information for classification. Besides, single patches vary in discriminative power. Forehead (F) and Nose (N) alone perform the worst with AUC around 0.88. Left cheek (LC) and right cheek (RC) perform much better, often above 0.93-0.98, suggesting that cheeks contain more useful features regarding rosacea. It's also worth noting that full face is not always the best. Interestingly, the AUC for full face images is sometimes worse than specific patch combinations (e.g., F+N+LC at 1.00 AUC). This suggests that noise or irrelevant information in the full face may reduce performance compared to carefully chosen regions.
\subsubsection{Patch size comparison}
When comparing large, medium, and small patches, the results indicate that patch size has only a modest effect on overall classification performance. Large patches provide slightly higher AUC values in some configurations, likely because they capture more contextual information around the region of interest. However, medium and small patches still maintain strong discriminative power, with only marginal drops in performance. This demonstrates that even localized input can preserve key diagnostic cues for rosacea detection.

From a practical standpoint, the ability of smaller patches to achieve nearly comparable results is important for both privacy preservation and model interpretability. Smaller regions limit the exposure of unnecessary facial information, reducing privacy risks, while also allowing for more targeted analysis of disease-prone regions. Together, these findings suggest that although large patches and full-face inputs remain optimal for accuracy, smaller patches strike a useful balance between performance and practical deployment considerations.

\section{Conclusion}
In this study, we investigated the impact of various patch-based strategies on the automated detection of rosacea using a ResNet-18 classification framework. By systematically comparing different combinations and sizes of facial regions, including the forehead, nose, and cheeks, we demonstrated that patch-based approaches can offer performance comparable to or even better than full-face models. In particular, combinations involving the left and right cheeks with other regions consistently achieved the highest AUC scores, indicating their clinical relevance and discriminative power.

Our results also highlight that while larger patches capture more contextual information, smaller and medium-sized patches can still maintain strong diagnostic accuracy. This balance supports the dual goals of preserving patient privacy and enhancing model interpretability. The ability to isolate and emphasize clinically meaningful areas suggests that patch-based learning offers a promising path forward for robust, privacy-conscious dermatological AI systems.

Overall, this comparative analysis provides a foundation for designing lightweight, region-focused models that are both effective and respectful of patient data confidentiality. Future work could explore adaptive patch selection mechanisms and broader clinical validation to further refine these methods for real-world deployment.


\end{document}